\documentclass{article}
\usepackage{spconf,amsmath,graphicx,hyperref}
\usepackage{cite}
\usepackage{amsmath,amssymb,amsfonts}
\usepackage{algorithmic}
\usepackage{multirow}
\usepackage{makecell}
\usepackage{graphicx}
\usepackage{textcomp}
\usepackage{xcolor}
\usepackage{float} 

\makeatletter
\renewcommand\section{\@startsection{section}{1}{0.3pt}%
  {1\baselineskip plus 0.2\baselineskip minus 0.2\baselineskip}%
  {0.55\baselineskip}%
  {\normalfont\scshape\centering}}

\renewcommand\subsection{\@startsection{subsection}{2}{0pt}%
  {0.7\baselineskip}%
  {0.4\baselineskip}%
  {\normalfont\itshape}}

\renewcommand\subsubsection{\@startsection{subsubsection}{3}{0pt}%
  {0.5\baselineskip}%
  {0.3\baselineskip}%
  {\normalfont\itshape}}
\makeatother

\makeatletter
\let\origthebibliography\thebibliography
\let\endorigthebibliography\endthebibliography
\renewenvironment{thebibliography}[1]{%
  \origthebibliography{#1}%
  \small 
  \setlength{\itemsep}{0pt}%
  \setlength{\parsep}{0pt}%
  \setlength{\parskip}{0pt}%
}{%
  \endorigthebibliography
}
\makeatother

\ninept                  
\usepackage[final]{microtype}  

\def\BibTeX{{\rm B\kern-.05em{\sc i\kern-.025em b}\kern-.08em
    T\kern-.1667em\lower.7ex\hbox{E}\kern-.125emX}}

\title{GLoRIA: Gated Low-Rank Interpretable Adaptation for Dialectal ASR}
\name{Pouya Mehralian\,$^{1}$, Melissa Farasyn\,$^{2}$, Anne Breitbarth\,$^{2}$, Anne-Sophie Ghyselen\,$^{3}$, Hugo Van hamme\,$^{1}$
}
\address{
$^{1}$ESAT/PSI, KU Leuven, Belgium \\
$^{2}$\,${\Delta}$iaLing, Ghent University, Belgium \\
$^{3}$ GLiMS \& MULTPIPLES, Ghent University, Belgium  }
\usepackage{eso-pic}

\newcommand{\IEEEpreprintnotice}{%
  \AddToShipoutPictureFG*{%
    \AtPageLowerLeft{%
      \raisebox{1.6\baselineskip}{%
        \makebox[\paperwidth]{%
          \parbox{0.82\paperwidth}{\tiny
          \copyright\ 2026 IEEE. Personal use of this material is permitted.
          Permission from IEEE must be obtained for all other uses, in any current
          or future media, including reprinting/republishing this material for
          advertising or promotional purposes, creating new collective works, for
          resale or redistribution to servers or lists, or reuse of any copyrighted
          component of this work in other works.}%
        }%
      }%
    }%
  }%
}
\AtBeginDocument{\IEEEpreprintnotice}

\begin{document}
%
\maketitle
\begin{abstract}
Automatic Speech Recognition (ASR) in dialect-heavy settings remains challenging due to strong regional variation and limited labeled data. We propose GLoRIA, a parameter-efficient adaptation framework that leverages metadata (e.g., coordinates) to modulate low-rank updates in a pre-trained encoder. GLoRIA injects low-rank matrices into each feed-forward layer, with a gating MLP determining the non-negative contribution of each LoRA rank-1 component based on location metadata. On the GCND corpus, GLoRIA outperforms geo-conditioned full fine-tuning, LoRA, and both dialect-specific and unified full fine-tuning, achieving state-of-the-art word error rates while updating under 10\% of parameters. GLoRIA also generalizes well to unseen dialects, including in extrapolation scenarios, and enables interpretable adaptation patterns that can be visualized geospatially. These results show metadata-gated low-rank adaptation is an effective, interpretable, and efficient solution for dialectal ASR.
\end{abstract}

\vspace{2pt}

\section{Introduction}
\label{sec:intro}
Automatic Speech Recognition (ASR) has advanced rapidly with deep learning and large pre-trained models, yet performance on dialectal speech remains limited due to phonetic, lexical, and syntactic variation combined with scarce annotated data. A promising direction is the use of auxiliary metadata that reflects linguistic variation. Geographical information is particularly effective and interpretable, as regional speech patterns often align with location~\cite{trudgill1999dialects,chambers2003sociolinguistic}. Prior work shows that leveraging utterance coordinates enables continuous interpolation between dialects and improved generalization compared to discrete labels~\cite{mehralian25_interspeech}. Geographic location has also been explored for written dialect translation \cite{simons-etal-2024-highly}.

Sociolinguistic studies show that core dialectal features often persist after migration, especially if acquired in childhood. This makes location a reliable proxy for dialect identity, with evidence from English~\cite{barker2016does}, German~\cite{schwarz2020reduction}, Swiss German~\cite{jeszenszky2024effects}, Spanish~\cite{Erker_Reffel_2021}, Portuguese~\cite{oushiro2020contrasting}, and Punjabi~\cite{sharma2011cognitive}. These findings further motivate ASR approaches that directly exploit geographical metadata in dialect-rich settings.

In this work, we propose \textbf{GLoRIA} (\textit{Gated Low-Rank Interpretable Adaptation}), a parameter-efficient framework that extends LoRA~\cite{lora} by modulating low-rank adaptation directions with coordinate-driven gating. GLoRIA offers three advantages: (i) \textit{Parameter efficiency}—only a fraction of parameters are updated, (ii) \textit{Empirical gains}—it consistently outperforms baselines including standard LoRA and full fine-tuning, and (iii) \textit{Interpretability}—its adaptations are geographically grounded and analyzable. Experiments on Southern Dutch dialects show state-of-the-art performance on both seen and unseen varieties, advancing dialect-aware ASR through efficient and interpretable location-aware adaptation.

\section{Related Work}

Various strategies have been developed to address ASR for dialect-heavy languages, focusing on model adaptation, the integration of metadata, and multi-task learning.

\subsection{Dialect-Specific Approaches}
A simple strategy is to train \textit{dialect-specific ASR models}, where each dialect has its own model. While effective, this requires maintaining multiple models and adds high computational cost~\cite{yoo2019adaptive,dorn2019dialect}.

\subsection{Metadata-Aware Unified Models}
To reduce complexity, unified models integrate dialect metadata, typically as categorical labels. Strategies include:  
(i) conditioning sequence-to-sequence models on dialect labels~\cite{toyama2024adapting,imaizumi2020dialect};  
(ii) injecting one-hot or learned label embeddings into encoder/decoder layers~\cite{li2018multi};  
(iii) mixture-of-experts architectures that route inputs to dialect-specific experts~\cite{zhou2025dialectmoe};  
(iv) hierarchical models with shared backbones and dialect-adaptive layers~\cite{yoo2019adaptive}; and  
(v) multi-task setups combining ASR with dialect classification~\cite{imaizumi2020dialect,imaizumi2022multi}.

\subsection{Continuous Geo-Conditioning}
Most prior work rely on discrete dialect labels, limiting generalization in regions where variation is gradual. Recent work shows that leveraging continuous geographical metadata  enables smooth interpolation across dialects and improved robustness to unseen varieties~\cite{mehralian25_interspeech}. Our approach follows this line by conditioning adaptation on continuous geo-information rather than categorical labels.

\section{Dataset}
All experiments are conducted on the GCND corpus~\cite{GCND_main}, which contains 411 hours of spontaneous Dutch dialect speech from Belgium, the southern Netherlands, and French Flanders. Each segment is transcribed and paired with precise geographical coordinates of the interview location. Participants were chosen for their strong local rootedness, having spent most of their lives in their villages, so the speech captures authentic and regionally stable dialects. The recordings feature an average of 3.8 speakers per session and cover a broad range of informal conversational topics. This makes GCND particularly well suited for geographically grounded ASR adaptation. 


\section{Pretrained Model}
The models utilized in this study are based on the Cascaded Encoder Dual Features architecture for ASR and subtitling~\cite{poncelet2023learning,poncelet2025leveragingbroadcastmediasubtitle}. 
The architecture uses two cascaded encoders: an ASR encoder for raw audio and a subtitle encoder for refined representations. Two Multi-Transformer decoders~\cite{multitransformer2021searchable} attend to both encoders in parallel, producing both verbatim and subtitle transcriptions.
Training employs a composite loss of weighted cross-entropy and CTC~\cite{CTC,hybridCTC}. In this study, we use the subtitle decoder output, as it is trained on far more data (14k vs. 300 hours for the ASR decoder) and dialect transcriptions are not strictly verbatim, making the subtitle decoder a more suitable choice. This Dutch-only model has 180M parameters and, in our experiments, outperforms Whisper large-v3~\cite{whisper} and OWSM-CTC-V4 1B~\cite{owsm_v4} on dialectal Dutch despite its smaller size. 

\vspace{4pt}

\section{Methods}

This section begins with a brief recap of the standard LoRA formulation, followed by the GLoRIA extension, which incorporates a coordinate-conditioned gating mechanism and regularization.

Let $\mathbf{W} \in \mathbb{R}^{d_{\text{out}}\times d_{\text{in}}}$ denote a weight matrix (e.g., from a feed-forward layer in a Conformer block \cite{gulati2020conformer}) of a pre-trained encoder–decoder ASR model, which remains frozen during adaptation. Both LoRA and GLoRIA perform a low-rank adaptation of $\mathbf{W}$ using two learnable matrices $\mathbf{A} \in \mathbb{R}^{d_{\text{out}} \times r}$ and $\mathbf{B} \in \mathbb{R}^{r \times d_{\text{in}}}$, where the rank $r \ll \min\{d_{\text{in}}, d_{\text{out}}\}$ controls the number of trainable parameters.

\vspace{3pt}

\subsection{LoRA Recap and GLoRIA}

LoRA (Low-Rank Adaptation) modifies the weight matrix by adding a low-rank update:
\vspace{-3pt}
\begin{equation}
    \mathbf{W}' = \mathbf{W} + \mathbf{A} \mathbf{B}
\end{equation}
\vspace{-12pt}

Only $\mathbf{A}$ and $\mathbf{B}$ (and optionally layer-norm parameters) are updated during fine-tuning.

GLoRIA extends LoRA by introducing a coordinate-conditioned gating mechanism via a diagonal matrix $\mathbf{E} \in \mathbb{R}^{r \times r}$:
\vspace{-3pt}
\begin{equation}
\begin{aligned}
\mathbf{W}' &= \mathbf{W} + \mathbf{A} \mathbf{E} \mathbf{B} 
            &= \mathbf{W} + \sum_{i=1}^r \gamma_i\, \mathbf{a}_i \mathbf{b}_i^\top
\end{aligned}
\end{equation}
\vspace{-6pt}

Here, $\gamma_i$ is the $i$-th diagonal entry of $\mathbf{E}$ (output of the gate-mlp), and $\mathbf{a}_i$ and $\mathbf{b}_i$ are the $i$-th columns of $\mathbf{A}$ and rows of $\mathbf{B}$, respectively. This formulation highlights that GLoRIA efficiently modulates each adaptation direction independently, with the final update being a sum of $r$ rank-1 components weighted by metadata-conditioned gates.




The diagonal elements of $\mathbf{E}$, denoted $\boldsymbol{\gamma} \in \mathbb{R}^{r}_{\geq 0}$, are predicted by a small neural network (\textit{gate-mlp}) that conditions on the recording's geographical coordinates $\mathbf{c} = (\mathrm{lat}, \mathrm{lng})$:
\vspace{1pt}
\begin{align}
    \mathbf{\boldsymbol{\gamma}} &= \mathrm{Softplus}(gate-mlp(c)) 
\end{align}
\vspace{-9pt}

Here, gate-mlp can be any multi-layer perceptron.
The Softplus activation ensures all gating values are non-negative, supporting interpretability by restricting each adaptation direction to contribute additively. This choice for the activation assumes the base model is relatively neutral with respect to dialects, so adaptation components are combined to introduce relevant dialectal traits. However, if the pre-trained model is already biased toward certain dialects, allowing negative gating values could be beneficial, as it would enable the model to suppress or subtract adaptation directions that are not appropriate for a given dialect. In such cases, alternative activations that permit signed values may be preferred.

\subsection{Regularization Losses}

To encourage the low-rank directions to be distinct and the gating to be selective, we add two additional regularizers:
\begin{itemize}
    \item \textbf{Orthonormality Loss:} Enforces diversity in adaptation directions by penalizing non-orthogonal columns in $\mathbf{A}$ and $\mathbf{B}$:
    \vspace{2pt}
    \begin{equation}
        \mathcal{L}_{\mathrm{orth}} = \| \mathbf{A}^\top \mathbf{A} - \mathbf{I} \|_F^2 + \| \mathbf{B} \mathbf{B}^\top - \mathbf{I} \|_F^2
    \end{equation}
    \vspace{-12pt}
    \item \textbf{Sparsity Loss:} To encourage $\gamma$ to focus on a small subset of adaptation components, we penalize its entropy. The entropy is normalized to maintain a comparable scale across different ranks:
    \vspace{-6pt}
    \begin{align}
    p_j &= \frac{\gamma_j}{\sum_{k=1}^r \gamma_k}, 
    &
    \mathcal{L}_{\mathrm{sp}}
    &= -\frac{1}{\log r}\sum_{j=1}^{r} p_j \log p_j.
\end{align}
\end{itemize}

\vspace{-12pt}

\subsection{Training Objective and Procedure}

The adaptation is applied to each feed-forward (FF) sub-layer in all encoder layers. For conformer architectures, which feature multiple FF blocks per layer (e.g., macaron and standard FFs), each block is adapted with separate low-rank matrices and gate-mlps. During adaptation, all backbone model parameters are frozen except for the GLoRIA parameters ($\mathbf{A}$, $\mathbf{B}$, gate-mlp and optionally layer normalizations). Training minimizes the ASR loss on the target data, with additional regularization:
\vspace{-3pt}
\begin{equation}
    \mathcal{L}_{\mathrm{total}} = \mathcal{L}_{\mathrm{ASR}} + \lambda_{\mathrm{orth}} \mathcal{L}_{\mathrm{orth}} + \lambda_{\mathrm{sp}} \mathcal{L}_{\mathrm{sp}}
\end{equation}
\vspace{-7pt}
where $\lambda_{\mathrm{orth}}$ and $\lambda_{\mathrm{sp}}$ control the strength of regularization terms.

\begin{table*}[t]
    \vspace{-15pt}
    \centering
    \caption{Word Error Rate (\%) comparison between GLoRIA and baseline models across different dialectal regions and models. The regions above the horizontal separator contain dialects that were seen during training, while the regions below unseen. For each region, the first column lists the lowest WER obtained among the five dialect-specific models trained on the training regions. It is followed by the joint model and the location-aware models, and then inference with the largest/latest Whisper and OWSM models, as well as the pre-trained model}
    \label{tab:main-results}
    \renewcommand{\arraystretch}{1.2}
    \scriptsize
    \setlength{\tabcolsep}{4pt}
    \begin{tabular}{l|c|c|cc|c|ccc|c}
    \hline
    \multicolumn{1}{c|}{\textit{Description}} 
      & \multicolumn{2}{c|}{\textit{Full Fine-Tuning, No Metadata}} 
      & \multicolumn{2}{c|}{\textit{Geo-Conditioned Full Fine-Tuning}} 
      & \multicolumn{1}{c|}{\textit{Geo-Conditioned}} 
      & \multicolumn{3}{c|}{\textit{General-Purpose Models}}
      & \multicolumn{1}{c}{} \\
    \hline
    \multirow{2}{*}{\textbf{Dialect Region}} 
      & \textbf{Dialect-} & \multirow{2}{*}{\makecell{\textbf{Joint}\\\textbf{Model}}}
      & \multirow{2}{*}{\makecell{\textbf{Coordinate}\\\textbf{Embedding}}} 
      & \multirow{2}{*}{\makecell{\textbf{Modified}\\\textbf{Feedforward}}} 
      & \multirow{2}{*}{\makecell{\textbf{GLoRIA}\\\textbf{(rank 128)}}} 
      & \textbf{OWSM} & \textbf{Whisper} & \textbf{Pre-trained} 
      & \multirow{2}{*}{\makecell{\textbf{Amount of}\\ \textbf{Test Data}}} \\
      & \textbf{Specific (Best)} & & & & & \textbf{V4-1B} & \textbf{Large-V3} & \textbf{Model} & \\
    \hline
    \textbf{Brabants} & 30.02 & 28.67 & 28.70 & 27.44 & \textbf{27.10} & 78.32 & 71.69 & 56.52 & 7h 47m \\
    \textbf{Frans-Vlaams} & 45.13 & 46.01 & 44.84 & 42.67 & \textbf{40.84} & 83.65 & 75.11 & 75.38 & 2h 34m \\
    \textbf{Oost-Vlaams} & 32.42 & 32.83 & 33.04 & 31.56 & \textbf{30.16} & 80.54 & 68.30 & 66.22 & 5h 03m \\
    \textbf{Oost-Vlaams$>$Brabants} & 35.89 & 32.65 & 33.21 & 31.98 & \textbf{30.28} & 75.34 & 68.04 & 64.51 & 3h 16m \\
    \textbf{West-Vlaams} & 29.53 & 28.61 & 29.17 & 27.36 & \textbf{26.70} & 76.25 & 62.91 & 58.95 & 8h 16m \\
    \hline
    \textbf{Limburgs} & \textbf{47.86} & 48.38 & 52.25 & 50.27 & 49.41 & 76.78 & 68.56 & 61.74 & 2h 44m \\
    \textbf{Limburgs$>$Brabants} & 38.74 & 39.28 & 41.08 & 38.78 & \textbf{37.61} & 76.01 & 62.07 & 58.32 & 1h 42m \\
    \textbf{Vlaams$>$Zeeuws} & 41.96 & 37.60 & 41.55 & 38.52 & \textbf{37.17} & 75.43 & 61.37 & 58.69 & 2h 41m \\
    \textbf{West-Vlaams$>$Oost-Vlaams} & 36.52 & 34.01 & 35.10 & 32.60 & \textbf{32.09} & 83.61 & 71.60 & 66.37 & 2h 46m \\
    \hline
\textbf{Average} & 37.56 & 36.45 & 37.66 & 35.69 & \textbf{34.59} & 78.44 & 67.74 & 62.97 & -- \\
\hline

    \end{tabular}
\end{table*}


\vspace{8pt}

\section{Experimental Setup}

\subsection{Dataset and Dialect Selection}
The dataset covers nine dialectal regions. Regions with $\geq$50 recordings are split 80/10/10 into train/validation/test, while those with fewer are excluded from training and contribute 10\% of their data to test only. Training dialects are Brabants, Frans-Vlaams, Oost-Vlaams, Oost-Vlaams$>$Brabants, and West-Vlaams; test-only dialects are Limburgs, Limburgs$>$Brabants, Vlaams$>$Zeeuws, and West-Vlaams$>$Oost-Vlaams. Transitional varieties (marked $>$) occur between core dialect regions and blend neighboring features, while still retaining distinct local features.

\subsection{Model Configuration}
Recordings are resampled to 16 kHz and represented with 80-dim mel-filterbanks plus 3 pitch features (25 ms window, 10 ms shift), concatenated and utterance-normalized. SpecAugment~\cite{park2019specaugment} is applied during training. Models are implemented in ESPnet~\cite{watanabe2018espnet} with a Conformer encoder preceded by 6-layer Conv2d subsampling and relative positional encodings. The encoder has 12 Conformer blocks (Swish activation, 8 heads, $d_{ff}=2048$, kernel size 31, $d_{model}=512$). The decoder is a 6-layer Transformer (8 heads, $d_{ff}=2048$, dropout 0.1). A lightweight subtitle encoder with 2 Transformer layers shares the same configuration.

In the GLoRIA models, the gate-mlps are two-layer multilayer perceptrons with a single hidden layer of size 32. These MLPs use the GeLU activation function in the hidden layer and a Softplus activation at the output to ensure smooth, positive gating values. The $\lambda_{\mathrm{sp}}$ and $\lambda_{\mathrm{orth}}$ are 5.0 and 0.8 respectively.

\subsection{Training Configuration}
Training uses gradient accumulation of 128. Full fine-tuning and LoRA models run for 100 epochs, while GLoRIA is trained for 40. Before adaptation, the pretrained model is dialect-finetuned for 5 epochs without metadata. New parameters (including embeddings) use Xavier uniform initialization, while gate-mlps are initialized as zero. Optimization is with Adam (lr=0.001) and WarmupLR (1500 warmup steps). CTC loss is excluded for the subtitle decoder.

\section{Results}

\subsection{Overall ASR Performance}

We evaluate GLoRIA against several conventional and geo-conditioned full fine-tuning baselines. These include (1) standard dialect-specific and joint fine-tuning approaches that rely solely on speech features, and (2) more sophisticated methods that incorporate geographical metadata during fine-tuning. Below, we describe these baselines in detail.

\subsubsection{Standard Fine-Tuning Baselines}
\begin{itemize}
    \item \textbf{Dialect-Specific Fine-Tuning:}  
Separate models fully fine-tuned on each dialect’s data.  

\item \textbf{Joint Fine-Tuning:}  
A single model fine-tuned on pooled dialectal data.  

Both rely solely on acoustic input and do not use geographical metadata.

\end{itemize}

\subsubsection{Geo-Conditioned Full Fine-Tuning Baselines}
We also compare to the coordinate-based methods of~\cite{mehralian25_interspeech}:  
\begin{itemize}
    \item \textbf{Coordinate Embedding:}  
    Longitude/latitude are mapped to $d_{model}$ embeddings and concatenated with speech features after subsampling, with a learnable \texttt{<coord>} token appended.  

    \item \textbf{Feedforward Modification:}  
    Two neurons encoding the scaled coordinates are injected into both feedforward blocks of each Conformer layer, enabling location-aware transformations without altering overall shape.  
\end{itemize}

Table~\ref{tab:main-results} summarizes the WERs achieved by GLoRIA compared to these baselines. GLoRIA consistently delivers the best accuracy across all seen dialects, surpassing both standard and geo-conditioned full fine-tuning methods: it outperforms the joint and coordinate-embedding models by 2.8\% and the modified feedforward model by 1.2\%, demonstrating the benefit of leveraging geographical metadata for in-distribution adaptation.

For held-out dialects, GLoRIA shows strong generalization, achieving the best WER in three of four unseen regions. In three of these cases, the model extrapolates to peripheral dialects beyond the training range, outperforming all baselines in two instances. This advantage likely reflects GLoRIA’s ability to extend adaptation directions from neighboring, linguistically similar regions. By contrast, geo-conditioned full fine-tuning degrades sharply outside the training convex hull. These results underscore the robustness of GLoRIA’s gating and low-rank adaptation, which generalize to both in- and out-of-distribution dialects by leveraging geographic structure.

Although WERs in this study can reach 40\%, this reflects the inherent difficulty of the GCND corpus, which includes far speech, variable recording quality, and the lack of a standardized transcription system for dialects. Table~\ref{tab:main-results} shows that even two of the largest recent models—Whisper Large V3 and OWSM\_CTC\_V4 1B—obtain WERs of 67.74 and 78.44, underscoring the challenge. Despite these high error rates, the GLoRIA outputs remain highly useful for native listeners—including those less familiar with the dialects—and provide considerable linguistic value beyond the raw scores.

\vspace{-1pt}

\subsection{GLoRIA vs. LoRA}

\vspace{-1pt}

Table~\ref{tab:gloria-lora} compares GLoRIA with standard LoRA adaptation across multiple ranks. Since LoRA is not conditioned on geographic metadata, its expressivity is limited when dialects are acoustically distant: the gap of $\sim$4\% WER between LoRA and the fully fine-tuned joint model highlights that low-rank adaptation alone cannot capture the variation across dialect regions.

In contrast, GLoRIA not only closes this gap but also surpasses the geo-conditioned full fine-tuning baselines. This improvement stems from its ability to exploit geographical information to interpolate and extrapolate across neighboring dialect regions, effectively reusing adaptation directions that align with linguistic similarity. Thus, while LoRA suffers from reduced expressivity, GLoRIA leverages structured location cues to achieve both parameter efficiency and superior recognition accuracy.
\vspace{-4pt}
\begin{table}[h]
    \vspace{-2.5pt}
    \centering
    \caption{WER (\%) comparison of LoRA and GLoRIA across ranks. The final row shows the fraction of model parameters updated.}
    \renewcommand{\arraystretch}{1.2}
    \scriptsize
    \setlength{\tabcolsep}{1.6pt}
    \label{tab:gloria-lora}
    \begin{tabular}{l|cc|cc|cc}
    \hline
    \textit{Description} & \multicolumn{2}{c|}{\textit{Rank 32}} & \multicolumn{2}{c|}{\textit{Rank 64}} & \multicolumn{2}{c}{\textit{Rank 128}} \\
    \hline
    \textbf{Dialect Region} & \textbf{LoRA} & \textbf{GLoRIA} & \textbf{LoRA} & \textbf{GLoRIA} & \textbf{LoRA} & \textbf{GLoRIA} \\
    \hline
    \textbf{Brabants} & 33.24 & 28.03 & 32.83 & 27.66 & 32.20 & \textbf{27.10} \\
    \textbf{Frans-Vlaams} & 55.90 & 43.04 & 54.31 & 42.53 & 50.48 & \textbf{40.84} \\
    \textbf{Oost-Vlaams} & 39.46 & 32.51 & 37.55 & 32.04 & 36.83 & \textbf{30.16} \\
    \textbf{Oost-Vlaams$>$Brabants} & 39.06 & 32.91 & 37.63 & 32.13 & 36.66 & \textbf{30.28} \\
    \textbf{West-Vlaams} & 35.18 & 27.70 & 34.48 & 27.40 & 32.46 & \textbf{26.70} \\
    \hline
    \textbf{Limburgs} & 52.45 & 54.19 & 52.14 & 49.51 & 52.42 & \textbf{49.41} \\
    \textbf{Limburgs$>$Brabants} & 42.64 & 39.23 & 42.00 & 38.80 & 41.94 & \textbf{37.61} \\
    \textbf{Vlaams$>$Zeeuws} & 42.52 & 38.99 & 42.39 & 38.25 & 41.19 & \textbf{37.17} \\
    \textbf{West-Vlaams$>$Oost-Vlaams} & 41.93 & 34.12 & 40.52 & 33.89 & 39.03 & \textbf{32.09} \\
    \hline
\textbf{Average} & 42.49 & 36.75 & 41.54 & 35.80 & 40.36 & \textbf{34.59} \\
    \hline
    \textbf{Trainable percentage} & 2.7\% & 2.7\% & 5.2\% & 5.3\% & 9.9\% & 10.0\% \\
    \hline
    \end{tabular}
\end{table}

\begin{figure*}[t]
    \centering
    \includegraphics[width=0.245\textwidth]{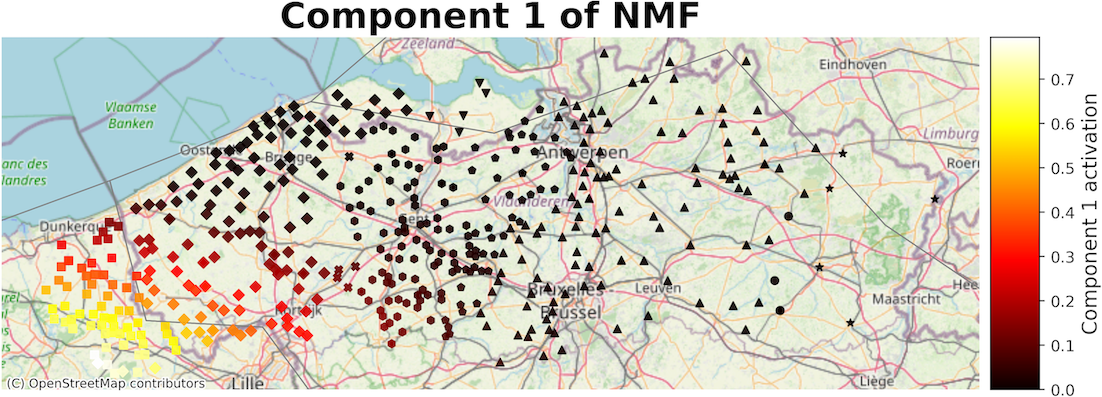}
    \includegraphics[width=0.245\textwidth]{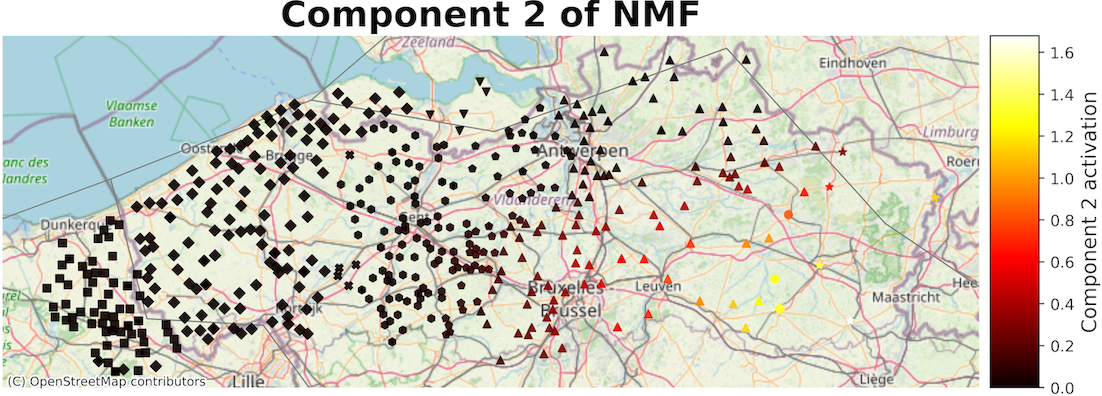}
    \includegraphics[width=0.245\textwidth]{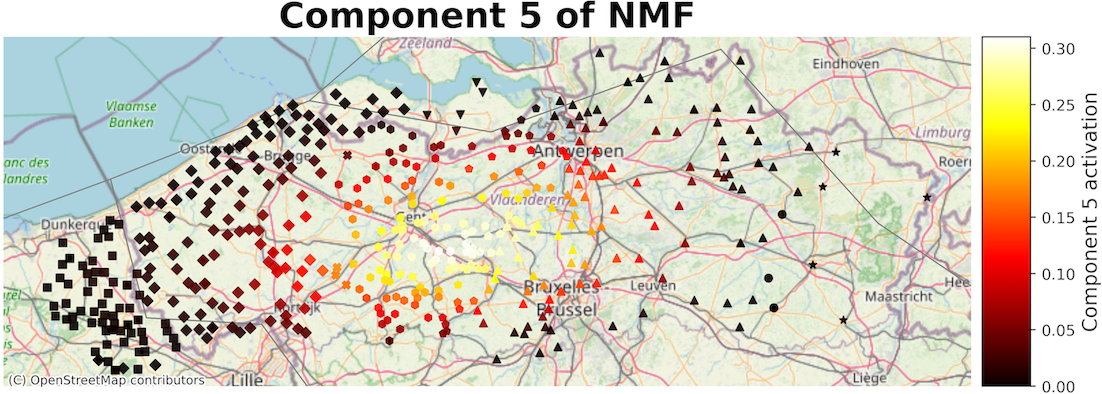}
    \includegraphics[width=0.245\textwidth]{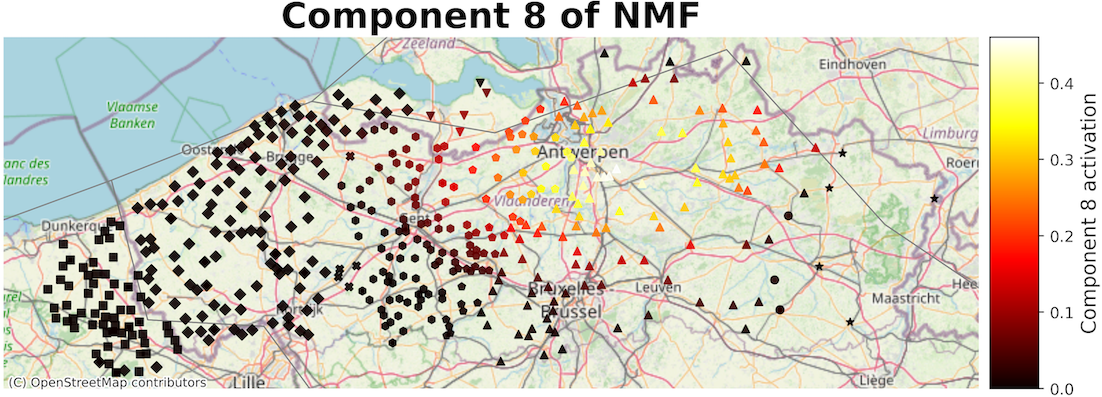}
    \\
    \includegraphics[width=0.95\textwidth]{legend_regions_horizontal.png}

    \vspace{-0.4em}

    \caption{\small Geographical distribution of activation for four NMF-derived adaptation components. Each point represents a location; color intensity reflects the degree of adaptation component usage at that location. The emergent spatial patterns closely correspond to known dialect regions. The patterns shown correspond to the Frans-Vlaams, Limburgs, Oost-Vlaams and Antwerp regions, from left to right.}
    \label{fig:region-maps}
\end{figure*}

\subsection{Interpreting Geographical Adaptation Directions via NMF}
\vspace{-1pt}
To assess the interpretability of the GLoRIA adaptation mechanism, we analyze how the gating vectors—produced by the coordinate-conditioned gate-mlp in each FF layer—vary across locations and dialect regions. 


\subsubsection{Extraction of Gating Activations}

For each recording location, we extract the output of every gate-mlp associated with each FF layer across the encoder. For a model configuration with 12 encoder layers, each containing two FF blocks (Macaron-style Conformer), and two FF layers per block, and with a GLoRIA adaptation rank of $r=128$, this yields a stacked gating vector of length $128 \times 48 = 6144$ for each location. With 488 unique locations, this process results in a $6144 \times 488$ matrix $\mathbf{G}$, where each column represents the complete adaptation signature for one location.

\vspace{4pt}
\begin{figure}[h]
  \centering
  \raisebox{-0.42\height}{\hbox{\includegraphics[width=0.042\textwidth,height=0.08\textheight]{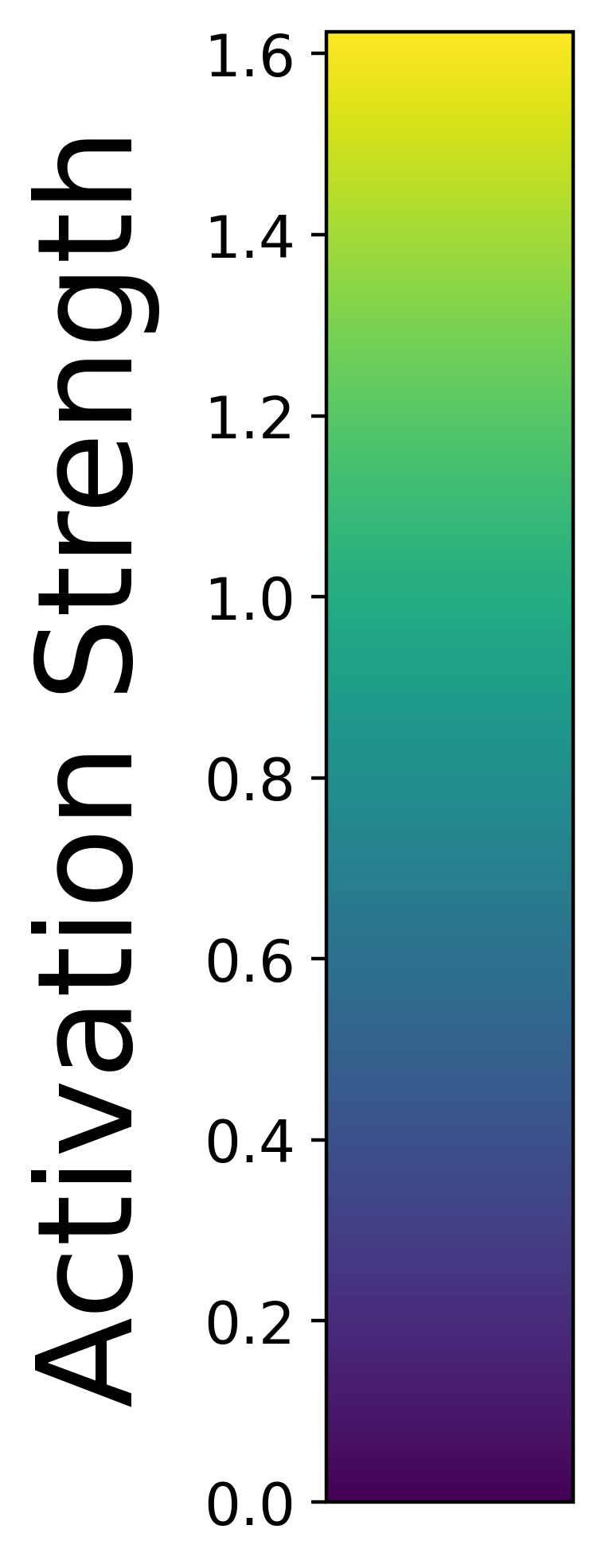}}}
  \hspace{0.81em}
  \raisebox{-.5\height}{\includegraphics[height=0.11\textheight]{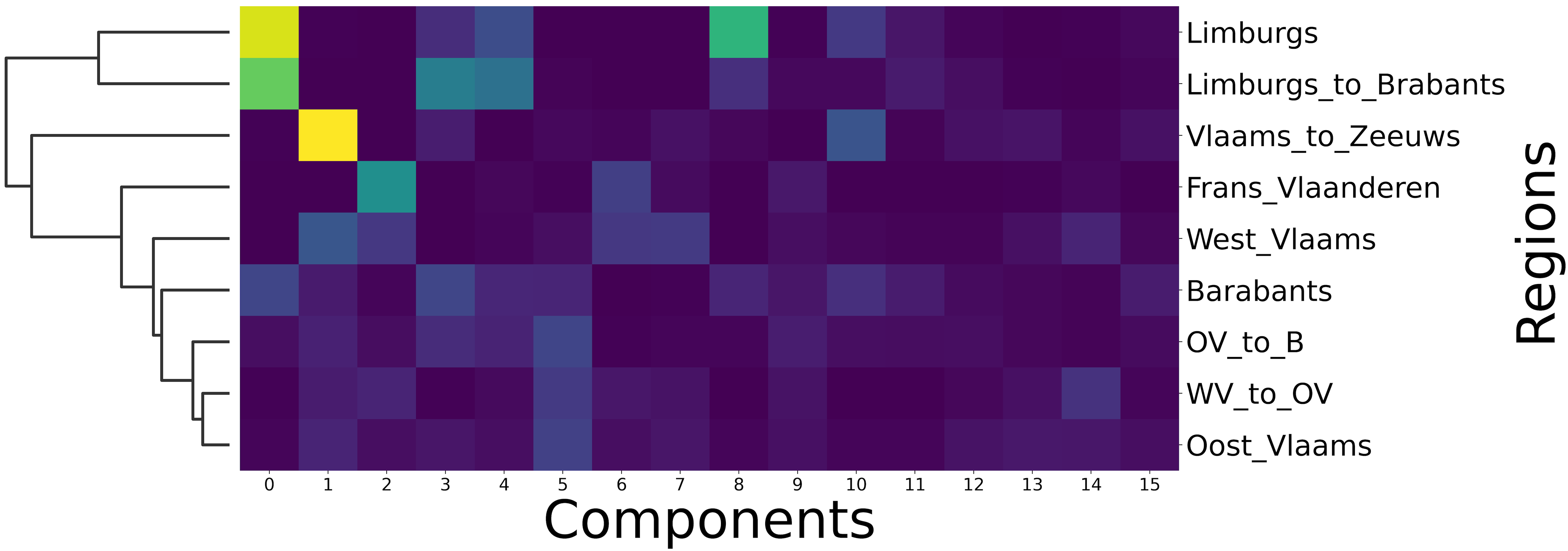}}
  \vspace{-0.4em}
  \caption{\small Clustered heatmap of NMF component activations averaged by dialect region. Regions that are geographically and acoustically closer cluster together, indicating interpretable adaptation behavior.}
  \label{fig:heatmap}
\end{figure}
\vspace{-6pt}
\subsubsection{Non-negative Matrix Factorization (NMF)}


Given the high dimensionality and redundancy in $\mathbf{G}$, we employ non-negative matrix factorization (NMF) to uncover the principal adaptation components across locations. We factorize $\mathbf{G} \approx \mathbf{S}\mathbf{L}$
where $\mathbf{S} \in \mathbb{R}_{\geq 0}^{6144 \times k}$ contains $k$ adaptation components (basis vectors), and $\mathbf{L} \in \mathbb{R}_{\geq 0}^{k \times 488}$ encodes the activation of each component per location. We use the Kullback-Leibler divergence as the distance metric for NMF and select $k=16$ based on the elbow point of the reconstruction loss curve. The factorization is performed using the multiplicative update (MU) solver \cite{lee2000algorithms} with non-negative double singular value decomposition averaging (NNDSVDA) initialization \cite{boutsidis2008svd}, and the algorithm runs for 3000 iterations to ensure convergence.

\subsubsection{Dialect Region Aggregation and Clustered Heatmap}


To relate adaptation patterns to dialect geography, we aggregate NMF activations by region. For each of the $k=16$ components, we compute the mean activation across all locations within the 9 dialect regions, yielding a $16 \times 9$ matrix. This matrix is visualized as a clustered heatmap (Fig.~\ref{fig:heatmap}), where columns represent components and rows correspond to regions. The clustering shows that geographically proximate and acoustically closer regions display comparable adaptation profiles. The figure also reveals region-selective activations, with only a subset of components contributing per region.


\subsubsection{Geospatial Visualization of Adaptation Components}



To illustrate the spatial structure captured by GLoRIA, we visualize the activation of four representative NMF components across the map. At each location, the activation values from $\mathbf{L}$ are mapped with color intensity proportional to component contribution. These geospatial maps (Fig.~\ref{fig:region-maps}) reveal clear patterns, with strong activations aligning to known dialectal boundaries—even though the model never observed dialect metadata during training. Moreover, maximum activation intensity for each region further indicates the relative degree of divergence from the standard language, with  Limburgs (component 2) showing stronger deviation than Oost-Vlaams (component 5).

Not all cities and villages within a dialect region are acoustically identical, and GLoRIA distinguishes these fine-grained differences, adapting continuously rather than being constrained by predefined boundaries. As a result, the model can partition dialects into more granular subregions when systematic differences arise, learning boundaries that best serve its adaptation objective. This is evident in Component 8 (rightmost map in Fig.~\ref{fig:region-maps}), where the model captures the known distinction of Antwerp within the Brabant dialect region—a difference also confirmed by prior acoustic studies \cite{Antwerp_DeSchutter, Antwerp_Vandekerckhove}.

\section{Discussion}

The results presented above highlight the strengths of GLoRIA in dialect-aware ASR. Through the use of coordinate-conditioned gating, GLoRIA identifies adaptation directions that are both effective for recognition and linguistically meaningful. The non-negative matrix factorization (NMF) analysis reveals that these directions are not arbitrary: the discovered components align with established dialectal regions and show clear spatial organization on the map.

While this study has focused on dialectal adaptation using geographical metadata, the GLoRIA methodology is broadly applicable to any domain where adaptation to structured metadata is beneficial. By conditioning the gating mechanism on alternative sources of information—such as age, channel characteristics, emotional state or socio-demographic variables—GLoRIA can dynamically adapt model behavior to a wide range of contextual cues. Future work will explore these avenues to further validate the generality and flexibility of GLoRIA, and to uncover new insights into interpretable, metadata-aware model adaptation.



\section{Conclusion}

An important contribution of this work is the demonstration that parameter-efficient adaptation need not come at the cost of interpretability, flexibility or performance. By modulating a small number of adaptation components according to metadata—in this case, geographic coordinates—GLoRIA provides a framework for fine-grained and interpretable adaptation. This approach moves the field closer to transparent and controllable ASR systems, with clear benefits for both practical deployment and linguistic analysis.

\vfill\pagebreak

\section{Acknowledgment}
This research was supported by the Flemish Government under ``Onderzoeksprogramma AI Vlaanderen''

{\small
\bibliographystyle{IEEEbib}
\bibliography{mybib}

@book{trudgill1999dialects,
  author = {Trudgill, Peter},
  title = {The Dialects of England},
  publisher = {Blackwell},
  year = {1999},
  edition = {2nd},
  address = {Oxford, UK}
}

@book{chambers2003sociolinguistic,
  author = {Chambers, J. K. and Trudgill, Peter},
  title = {Dialectology},
  publisher = {Cambridge University Press},
  year = {2003},
  edition = {2nd},
  address = {Cambridge, UK}
}

@inproceedings{yoo2019adaptive,
  author = {S. Yoo and I. Song and Y. Bengio},
  title = {A Highly Adaptive Acoustic Model for Accurate Multi-dialect Speech Recognition},
  booktitle = {Proc. ICASSP},
  year = {2019},
  pages = {5716--5720},
  doi = {10.1109/ICASSP.2019.8683705}
}

@inproceedings{li2018multi,
  author = {B. Li and T. N. Sainath and K. C. Sim and M. Bacchiani and E. Weinstein and P. Nguyen and Z. Chen and Y. Wu and K. Rao},
  title = {Multi-Dialect Speech Recognition with a Single Sequence-to-Sequence Model},
  booktitle = {Proc. ICASSP},
  year = {2018},
  pages = {4749--4753},
  doi = {10.1109/ICASSP.2018.8461886}
}

@inproceedings{imaizumi2020dialect,
  author = {R. Imaizumi and R. Masumura and S. Shiota and H. Kiya},
  title = {Dialect-aware modeling for end-to-end {Japanese} dialect speech recognition},
  booktitle = {Proc. APSIPA ASC},
  year = {2020},
  pages = {297--301}
}

@article{imaizumi2022multi,
  title={End-to-end {Japanese} multi-dialect speech recognition and dialect identification with multi-task learning},
  author={Imaizumi, R. and Masumura, R. and Shiota, S. and Kiya, H. and others},
  journal={APSIPA Transactions on Signal and Information Processing},
  volume={11},
  number={1},
  year={2022},
  publisher={Now Publishers, Inc.}
}

@inproceedings{zhou2025dialectmoe,
  title={DialectMoE: An End-to-End Multi-dialect Speech Recognition Model with Mixture-of-Experts},
  author={Zhou, J. and Gao, S. and Yu, Z. and Dong, L. and Wang, W.},
  booktitle={China National Conference on Chinese Computational Linguistics},
  pages={243--258},
  year={2024},
  organization={Springer}
}

@inproceedings{poncelet2023learning,
  title={Learning to Jointly Transcribe and Subtitle for End-To-End Spontaneous Speech Recognition},
  author={Poncelet, Jakob and Van Hamme, Hugo},
  booktitle={2022 IEEE Spoken Language Technology Workshop (SLT)},
  pages={182--189},
  year={2023},
  organization={IEEE}
}

@article{multitransformer2021searchable,
  title={Searchable hidden intermediates for end-to-end models of decomposable sequence tasks},
  author={Dalmia, Siddharth and Yan, Brian and Raunak, Vikas and Metze, Florian and Watanabe, Shinji},
  journal={arXiv preprint arXiv:2105.00573},
  year={2021}
}

@article{toyama2024adapting,
  title={Adapting Large-Scale Pre-trained Models for Unified Dialect Speech Recognition Model.},
  author={Toyama, T and Kai, A and Kamiya, Y and Takahashi, N},
  journal={Acta Physica Polonica: A},
  volume={146},
  number={4},
  year={2024}
}

@inproceedings{dorn2019dialect,
  title={Dialect-specific models for automatic speech recognition of {African American Vernacular English}},
  author={Dorn, Rachel},
  booktitle={Proceedings of the Student Research Workshop Associated with RANLP 2019},
  pages={16--20},
  year={2019}
}

@misc{poncelet2025leveragingbroadcastmediasubtitle,
      title={Leveraging Broadcast Media Subtitle Transcripts for Automatic Speech Recognition and Subtitling}, 
      author={Jakob Poncelet and Hugo Van hamme},
      year={2025},
      eprint={2502.03212},
      archivePrefix={arXiv},
      primaryClass={eess.AS},
      url={https://arxiv.org/abs/2502.03212}, 
}

@article{hybridCTC,
  author = {S. Watanabe and T. Hori and S. Kim and J. R. Hershey and T. Hayashi},
  title = {Hybrid {CTC/Attention} Architecture for End-to-End Speech Recognition},
  journal = {IEEE J. Sel. Top. Signal Process.},
  year = {2017},
  volume = {11},
  number = {8},
  pages = {1240--1253},
  doi = {10.1109/JSTSP.2017.2763455}
}

@inproceedings{CTC,
author = {Graves, Alex and Fern\'{a}ndez, Santiago and Gomez, Faustino and Schmidhuber, J\"{u}rgen},
title = {Connectionist temporal classification: labelling unsegmented sequence data with recurrent neural networks},
year = {2006},
isbn = {1595933832},
publisher = {Association for Computing Machinery},
address = {New York, NY, USA},
url = {https://doi.org/10.1145/1143844.1143891},
doi = {10.1145/1143844.1143891},
booktitle = {Proceedings of the 23rd International Conference on Machine Learning},
pages = {369–376},
numpages = {8},
location = {Pittsburgh, Pennsylvania, USA},
series = {ICML '06}
}

@inproceedings{whisper,
  title={Robust speech recognition via large-scale weak supervision},
  author={Radford, Alec and Kim, Jong Wook and Xu, Tao and Brockman, Greg and McLeavey, Christine and Sutskever, Ilya},
  booktitle={International conference on machine learning},
  pages={28492--28518},
  year={2023},
  organization={PMLR}
}

@article{GCND_main,
  title={Het gesproken corpus van de zuidelijk-Nederlandse dialecten},
  author={Breitbarth, Anne and Farasyn, Melissa and Ghyselen, Anne-Sophie and Van Keymeulen, Jacques},
  journal={Handelingen-Koninklijke Zuid-Nederlandse maatschappij voor taal-en letterkunde en geschiedenis},
  volume={72},
  year={2018},
  publisher={Koninklijke Zuid-Nederlandse Maatschappij voor Taal-en Letterkunde en~…}
}

@article{gulati2020conformer,
  title={Conformer: Convolution-augmented Transformer for Speech Recognition},
  author={Anmol Gulati and James Qin and Chung-Cheng Chiu and Niki Parmar and Yu Zhang and Jiahui Yu and Wei Han and Shibo Wang and Zhengdong Zhang and Yonghui Wu and Ruoming Pang},
  journal={ArXiv},
  year={2020},
  volume={abs/2005.08100},
  url={https://api.semanticscholar.org/CorpusID:218674528}
}

@inproceedings{watanabe2018espnet,
  author = {S. Watanabe and T. Hori and S. Karita and T. Hayashi and J. Nishitoba and Y. Unno and N. E. Yalta Soplin and J. Heymann and M. Wiesner and N. Chen and A. Renduchintala and T. Ochiai},
  title = {{ESPnet}: End-to-End Speech Processing Toolkit},
  booktitle = {Proc. Interspeech},
  year = {2018},
  pages = {2207--2211},
  doi = {10.21437/Interspeech.2018-1456}
}

@inproceedings{park2019specaugment,
  title={SpecAugment: A Simple Data Augmentation Method for Automatic Speech Recognition},
  author={Daniel S. Park and William Chan and Yu Zhang and Chung-Cheng Chiu and Barret Zoph and Ekin Dogus Cubuk and Quoc V. Le},
  booktitle={Interspeech},
  year={2019},
  url={https://api.semanticscholar.org/CorpusID:121321299}
}

@article{lora,
  title={Lora: Low-rank adaptation of large language models.},
  author={Hu, Edward J and Shen, Yelong and Wallis, Phillip and Allen-Zhu, Zeyuan and Li, Yuanzhi and Wang, Shean and Wang, Lu and Chen, Weizhu and others},
  journal={ICLR},
  volume={1},
  number={2},
  pages={3},
  year={2022}
}

@article{barker2016does,
  title={Does Geographic Relocation Induce the Loss of Features from a Single Speaker’s Native Dialect?},
  author={Barker, Hollie},
  journal={Lifespans and Styles},
  volume={2},
  number={1},
  pages={27--34},
  year={2016}
}

@article{sharma2011cognitive,
  author = {D. Sharma and L. Sankaran},
  title = {Cognitive and social forces in dialect shift: Gradual change in London Asian speech},
  journal = {Lang. Var. Change},
  year = {2011},
  volume = {23},
  number = {3},
  pages = {399--428}
}

@inbook{Erker_Reffel_2021, place={Cambridge}, title={The Persistence of Dialectal Differences in U.S. Spanish: /s/ Deletion in Boston and New York City}, booktitle={English and Spanish: World Languages in Interaction}, publisher={Cambridge University Press}, author={Erker, Daniel and Reffel, Madeline}, editor={Perez, Danae and Hundt, Marianne and Kabatek, Johannes and Schreier, DanielEditors}, year={2021}, pages={312–334}
}

@article{oushiro2020contrasting,
  title={Contrasting age of arrival and length of residence in dialect contact},
  author={Oushiro, Livia},
  year={2020}
}

@incollection{schwarz2020reduction,
  title={Reduction and persistence of phonological dialect features in {German}},
  author={Schwarz, Christian},
  booktitle={Intermediate language varieties},
  pages={103--124},
  year={2020},
  publisher={John Benjamins Publishing Company}
}

@article{jeszenszky2024effects,
  title={Effects of mobility on dialect change: Introducing the Linguistic Mobility Index},
  author={Jeszenszky, P{\'e}ter and Steiner, Carina and Leemann, Adrian},
  journal={Plos one},
  volume={19},
  number={4},
  pages={e0300735},
  year={2024},
  publisher={Public Library of Science San Francisco, CA USA}
}

@article{lee2000algorithms,
  title={Algorithms for non-negative matrix factorization},
  author={Lee, Daniel and Seung, H Sebastian},
  journal={Advances in neural information processing systems},
  volume={13},
  year={2000}
}

@article{boutsidis2008svd,
  title={SVD based initialization: A head start for nonnegative matrix factorization},
  author={Boutsidis, Christos and Gallopoulos, Efstratios},
  journal={Pattern recognition},
  volume={41},
  number={4},
  pages={1350--1362},
  year={2008},
  publisher={Elsevier}
}

@inproceedings{mehralian25_interspeech,
  title     = {{Leveraging Geographic Metadata for Dialect-Aware Speech Recognition}},
  author    = {Pouya Mehralian and Hugo {Van hamme}},
  year      = {2025},
  booktitle = {{Interspeech 2025}},
  pages     = {1153--1157},
  doi       = {10.21437/Interspeech.2025-1839},
  issn      = {2958-1796},
}

@inproceedings{simons-etal-2024-highly,
    title = "Highly Granular Dialect Normalization and Phonological Dialect Translation for {L}imburgish",
    author = "Simons, Andreas  and
      De Pascale, Stefano  and
      Franco, Karlien",
    booktitle = "Proceedings of the Eleventh Workshop on NLP for Similar Languages, Varieties, and Dialects (VarDial 2024)",
    month = jun,
    year = "2024",
    address = "Mexico City, Mexico",
    publisher = "Association for Computational Linguistics",
    url = "https://aclanthology.org/2024.vardial-1.13/",
    doi = "10.18653/v1/2024.vardial-1.13",
    pages = "152--162"
}

@inproceedings{owsm_v4,
  author = {Y. Peng and M. Shakeel and Y. Sudo and W. Chen and J. Tian and C.-J. Lin and S. Watanabe},
  title = {OWSM v4: Improving Open Whisper-Style Speech Models via Data Scaling and Cleaning},
  booktitle = {Proc. Interspeech},
  year = {2025},
  pages = {2225--2229},
  doi = {10.21437/Interspeech.2025-1062}
}

@article{Antwerp_Vandekerckhove,
url = {https://doi.org/10.1515/IJSL.2009.017},
title = {Dialect loss and dialect vitality in Flanders},
title = {},
author = {Reinhild Vandekerckhove},
pages = {73--97},
volume = {2009},
number = {196-197},
journal = {International Journal of the Sociology of Language},
doi = {doi:10.1515/IJSL.2009.017},
year = {2009},
lastchecked = {2025-09-15}
}

@inbook{Antwerp_DeSchutter,
url = {https://doi.org/10.1515/9783110261332.297},
title = {16. The dialects of Brabant: Grammatical properties},
booktitle = {Volume 3 Dutch},
author = {Georges De Schutter},
editor = {Frans Hinskens and Johan Taeldeman},
publisher = {De Gruyter Mouton},
address = {Berlin, Boston},
pages = {297--318},
doi = {doi:10.1515/9783110261332.297},
isbn = {9783110261332},
year = {2014},
lastchecked = {2025-09-15}
}
}
\end{document}